\pdfoutput=1

\documentclass[11pt]{article}

\usepackage[]{emnlp2021}

\usepackage{times}
\usepackage{latexsym}

\usepackage[T1]{fontenc}

\usepackage[utf8]{inputenc}

\usepackage{microtype}

\usepackage{amsmath}
\usepackage{mathtools}
\usepackage{amsthm}
\usepackage{amssymb}
\usepackage{enumitem}

\newcommand{\xx}{\mathbf{x}}

\newcommand{\hh}{\mathbf{h}}

\newcommand{\vv}{\mathbf{v}}

\newcommand{\HH}{\mathbf{H}}

\global\long\def\cF{\mathcal{F}}
\global\long\def\cS{\mathcal{S}}
\global\long\def\cG{\mathcal{G}}
\global\long\def\cP{\mathcal{P}}
\global\long\def\cT{\mathcal{T}}

\newcommand{\commentout}[1]{}

\usepackage{amsmath}
\usepackage{multirow}
\usepackage{booktabs}
\usepackage{graphicx}
\usepackage[normalem]{ulem}
\usepackage[cal=cm]{mathalfa}

\usepackage{amsthm}

\theoremstyle{definition}

\newcommand\jb[1]{\textcolor{red}{[JB: #1]}}

\newcommand\nl[1]{{\it``#1''}}

\newcommand\genbert{\textsc{GenBERT}}
\newcommand\msegenbert{\textsc{MseGenBERT}}
\newcommand\msebert{\textsc{MseBERT}}
\newcommand\reader{\textsc{Reader}}
\newcommand\readersum{\textsc{ReaderSum}}
\newcommand\drop{\textsc{DROP}}
\newcommand\hotpotqa{\textsc{HotpotQA}}
\newcommand\cnndaily{\textsc{CNN/DailyMail}}

\newcommand\omse{$o_{\textit{mse}}$}
\newcommand\ogen{$o_{\textit{gen}}$}
\newcommand\otype{$o_{\textit{type}}$}
\newcommand\osse{$o_{\textit{sse}}$}
\newcommand\osum{$o_{\textit{sum}}$}

\title{What's in Your Head?\\ Emergent Behaviour in Multi-Task Transformer Models}

\author{Mor Geva$^{1,2}$ ~~~~~ Uri Katz$^{1}$ ~~~~~ Aviv Ben-Arie$^{3}$ ~~~~~
Jonathan Berant$^{1,2}$ \\
$^1$Blavatnik School of Computer Science, Tel-Aviv University \\
$^2$Allen Institute for Artificial Intelligence \\
$^3$Independent Researcher \\
\small{\texttt{\{morgeva,urikatz1\}@mail.tau.ac.il}},  \small{\texttt{avivba@gmail.com}},  \small{\texttt{joberant@cs.tau.ac.il}}
}

\begin{document}
\maketitle

\begin{abstract}
The primary paradigm for multi-task training in natural language processing is to represent the input with a shared pre-trained language model, and add a small, thin network (\emph{head}) per task.
Given an input, a \emph{target head} is the head that is selected for outputting the final prediction.
In this work, we examine the behaviour of \emph{non-target heads}, that is, the output of heads when given input that belongs to a different task than the one they were trained for.
We find that non-target heads exhibit emergent behaviour, which may either explain the target task, or generalize beyond their original task. 
For example, in a numerical reasoning task, a span extraction head extracts from the input the arguments to a computation that results in a number generated by a target generative head. 
In addition, a summarization head that is trained with a target question answering head, outputs query-based summaries when given a question and a context from which the answer is to be extracted.
This emergent behaviour suggests that multi-task training leads to non-trivial extrapolation of skills, which can be harnessed for interpretability and generalization.
\end{abstract}

\section{Introduction}
\label{sec:introduction}

\begin{figure}[t]
    \centering
    \includegraphics[scale=0.4]{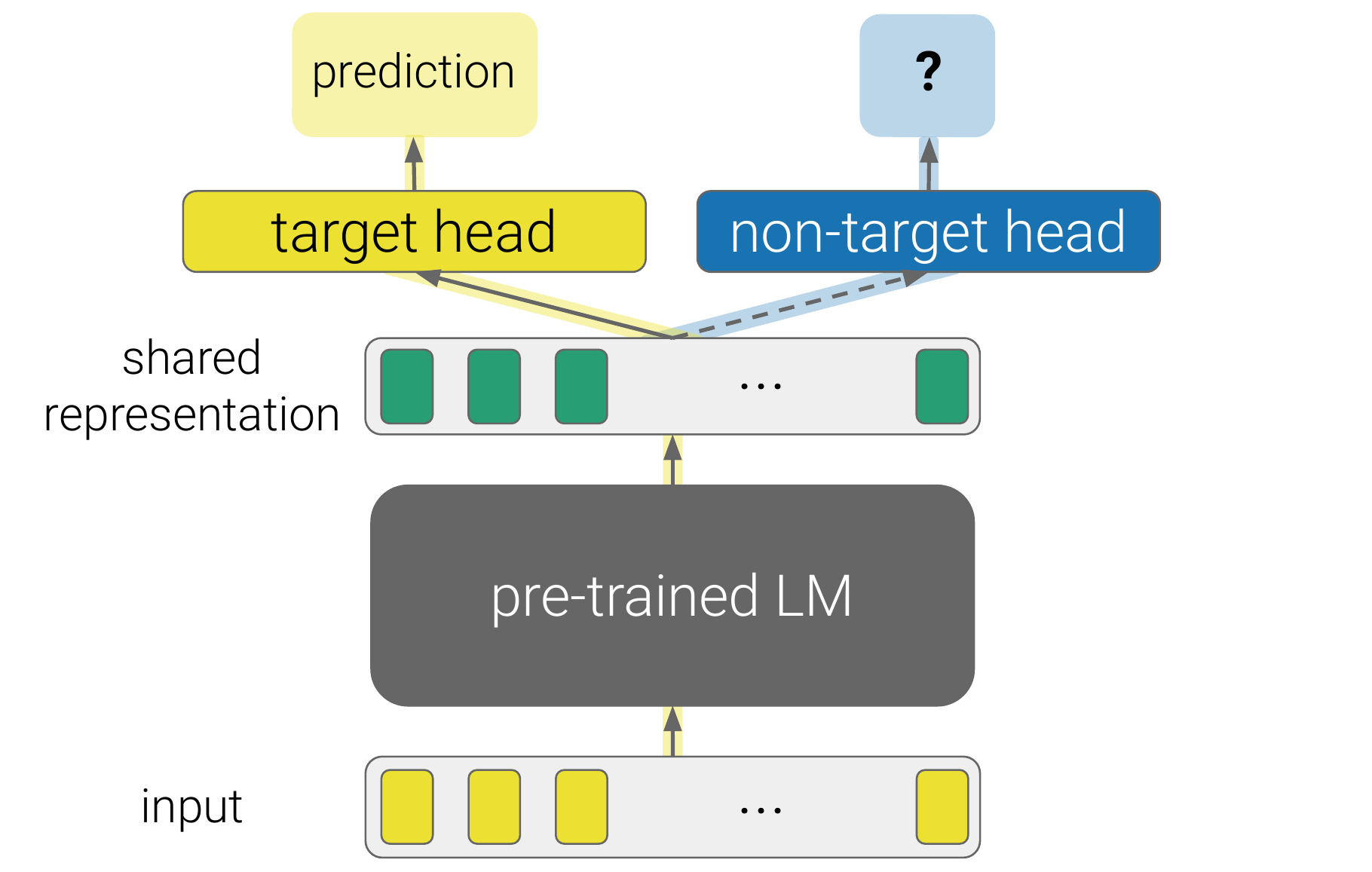}
    \caption{An illustration of multi-task  training with a pre-trained LM. Given an input for one of the tasks, a shared representation is computed with a pre-trained LM (green). The target head outputs the prediction, while the other heads are ignored. In this work, we characterize the behaviour of the \emph{non-target} head. 
    } 
    \label{figure:intro}
\end{figure}

The typical framework for training a model in natural language processing to perform multiple tasks is to have a shared pre-trained language model (LM), and add a small, compact neural network, often termed \emph{head}, on top of the LM, for each task \cite{clark2019bam, liu2019multi, nishida2019answering,hu2021transformer}.
The heads are trained in a supervised manner, each on labelled data collected for the task it performs \cite{devlin2018bert}. At inference time, the output is read out of a selected \emph{target head}, while the outputs from the other heads are discarded (Figure~\ref{figure:intro}).

What is the nature of predictions made by non-target heads given inputs directed to the target head?
One extreme possibility is that the pre-trained LM identifies the underlying task, and constructs unrelated representations for each task. In this case, the output of the non-target head might be arbitrary, as the non-target head observes inputs considerably different from those it was trained on. Conversely, the pre-trained LM might create similar representations for all tasks, which can lead to meaningful interactions between the heads.

In this work, we test whether such interactions occur in multi-task transformer models, and if non-target heads decode useful information given inputs directed to the target head.  
We show that multi-head training leads to a \emph{steering effect}, where the target head guides the behaviour of the non-target head, steering it to exhibit emergent behaviour, which can explain the target head's predictions, or generalize beyond the task the non-target head was trained for.


\begin{figure*}[t]
    \centering
    \includegraphics[scale=0.36]{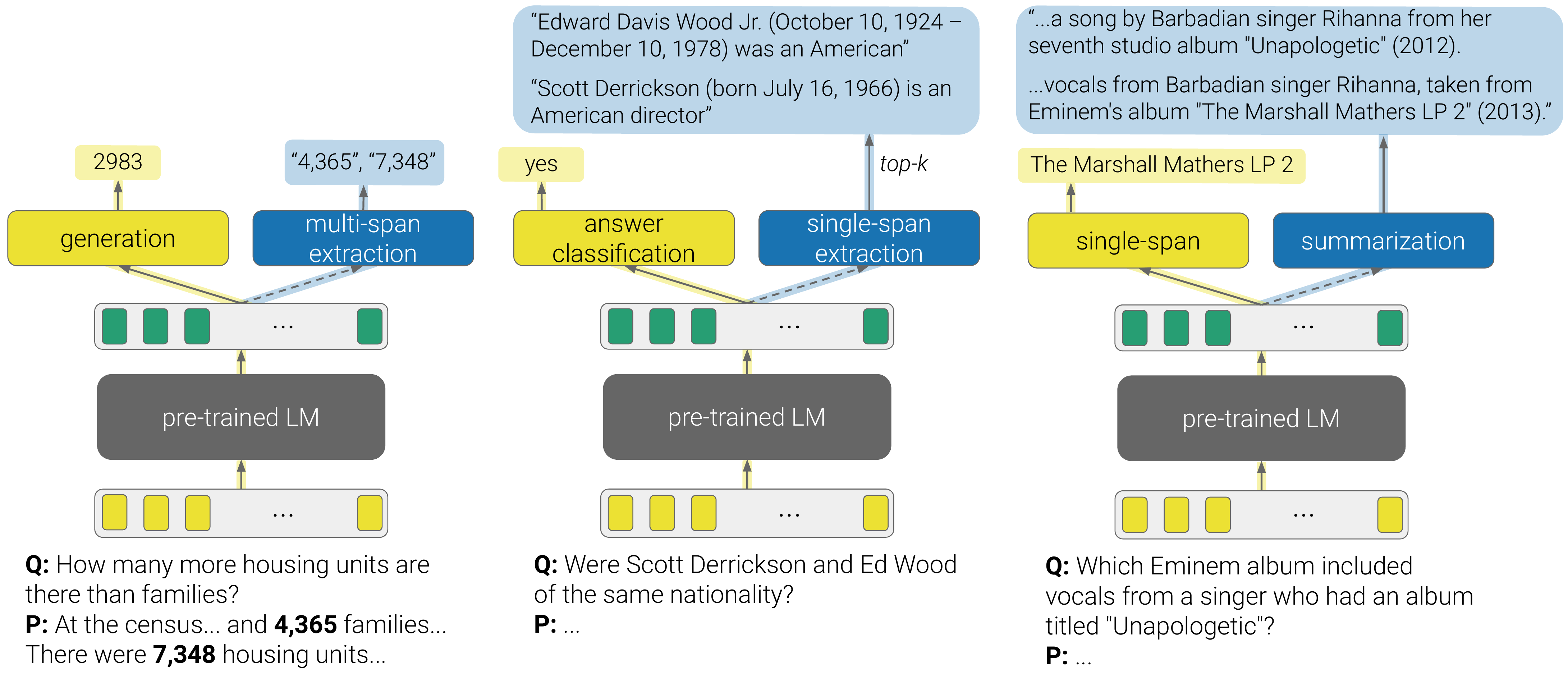}
    \caption{An overview of the three models analyzed in this work. For each model, the target head, which outputs the model's prediction, is shown on the left (in yellow). The non-target head, shown on the right (in blue) exhibits new behaviour \emph{without being trained for this objective}.}
    \label{figure:overview}
\end{figure*}

We study the ``steering effect'' in three multi-head models (Figure~\ref{figure:overview}).
In a numerical reading comprehension task \cite{dua2019drop}, the model is given a question and paragraph and either uses an extractive head to output an input span, or a generative head to generate a number using arithmetic operations over numbers in the input (Figure~\ref{figure:overview}, left). Treating the extractive head as the non-target head, we observe that it tends to output the arguments to the arithmetic operation 
performed by the decoder, and that successful argument extraction is correlated with higher performance. Moreover, we perform \emph{interventions} \cite{woodward2005making, elazar2021amnestic},
where we modify the representation based on the output of the extractive head, 
and show this leads to predictable changes in the behaviour of the generative head.
Thus, we can use the output of the non-target head to improve \emph{interpretability}.

We observe a similar phenomenon in multi-hop question answering (QA) model \cite{yang2018hotpotqa}, where a non-target span extraction head outputs supporting evidence for the answer predicted by a classification head (Figure~\ref{figure:overview}, center).
This emerging interpretability is considerably different from methods that \emph{explicitly train} models to output explanations \cite{perez2019finding, schuff2020f1}.



Beyond interpretability, we observe non-trivial extrapolation of skills when performing multi-task training on extractive summarization \cite{hermann2015teaching} and multi-hop QA (Figure\ref{figure:overview}, right). Specifically, a head trained for extractive summarization outputs supporting evidence for the answer when given a question and a paragraph, showing that multi-task training steers its behaviour towards \emph{query-based summarization}. We show this does not happen in lieu of multi-task training.

To summarize, we investigate the behaviour of non-target heads in three multi-task transformer 
models, and find that without any dedicated training, non-target heads provide explanations for the predictions of target heads, and exhibit capabilities beyond the ones they were trained for. 
This extrapolation of skills can be harnessed for many applications. For example, teaching models new skills could by done by training on combinations of tasks different from the target task. This would be useful when labeled data is not available or hard to collect. Also, training an additional head that extracts information from the input could be applied as a general practice for model interpretability.

\begin{table*}[t]
    \centering
    \footnotesize
    \begin{tabular}{p{3.4cm}|p{3.2cm}|p{2.3cm}|p{5cm}}
         Target head ($o_t$) & Steered head ($o_s$) & Dataset(s) & Emergent behaviour \\ \toprule
         Generation & Multi-span extraction & \drop{} & $o_s$ extracts numerical arguments for $o_t$ \\ \hline
         Classification (yes/no/span/no-answer) & Single-span extraction & \hotpotqa{} & $o_s$ extracts supporting facts for $o_t$ \\ \hline
         Single-span extraction & Extractive summarization & \hotpotqa{}, \cnndaily{} & $o_s$  performs query-based summarization \\ \hline
    \end{tabular}
    \caption{A summary of the main findings in each of the settings investigated in this work. 
    }
    \label{table:settings}
\end{table*}

\section{Multi-Head Transformer Models}
\label{sec:setting}
The prevailing method for
training models to perform NLP tasks is to add parameter-thin heads on top of a pre-trained LM, and fine-tune the entire network on labeled examples \cite{devlin2018bert}. 

Given a text input with $n$ tokens $\xx = \langle x_1, \dots, x_n \rangle$, the model first computes contextualized representations $\HH = \langle \hh_1, \dots, \hh_n \rangle = LM_{\mathbf{\theta}}(\xx)$ using the pre-trained LM parameterized by $\mathbf{\theta}$. These representations are then fed into the output heads, with each head $o$ estimating the probability $p_{\psi_o}(y \mid \HH)$ of the true output $y$ given the encoded input $\HH$ and the parameters $\mathbf{\psi}_o$ of $o$. 
The head that produces the final model output, termed the \emph{target head}, is chosen either deterministically, based on the input task,
or predicted by an output head classifier $p(o \mid \xx)$.
Predictions made by non-target heads are typically ignored. When $p(o \mid \xx)$ is deterministic it can be viewed as an indicator function for the target head.

Training multi-head transformer models is done by
marginalizing over the set of output heads $\mathcal{O}$, and maximizing the probability
$$ p(y \mid \xx) = \sum_{o\in \mathcal{O}} p(o \mid \xx)\cdot p(y \mid \HH, \mathbf{\psi}_o), $$
where $p(y \mid \HH, \mathbf{\psi}_o)>0$ only if  $y$ is in the output space of the head $o$.

For a head $o$, we denote by $\cS_o$ the set of examples $(x,y)$ such that $y$ is in the output space of $o$, and by $\bar{\cS}_o$ the other training examples. The sets $\cS_o$ and $\bar{\cS}_o$ may consist of examples from different tasks (e.g., question answering and text summarization), or of examples from the same task but with different output formats (e.g., yes/no vs. span extraction questions).
Our goal is to evaluate the predictions of $o$ on examples from $\bar{\cS}_o$, for which another head $o'$ is the target head, and the relation between these outputs and the predictions of $o'$. 

In the next sections, we will show that the predictions of $o$ interact with those of $o'$. 
We will denote by $o_t$ the target head, and by $o_s$ the \emph{steered head}.




\section{Overview: Experiments \& Findings}
\label{sec:overview}

This section provides an overview of our experiments, which are discussed in detail in \S\ref{sec:numerical_arguments_extraction}, \S\ref{sec:supporting_facts_extraction}, \S\ref{sec:query_based_summarization}.


Given a model with a target head $o_t$ and a steered head $o_s$, our goal is to understand the behaviour of $o_s$ on inputs where $o_t$ provides the prediction. To this end, we focus on head combinations, where $o_s$ is expressive enough to explain the outputs of $o_t$, but unlike most prior work aiming to explain by examining model outputs \cite{perez2019finding, schuff2020f1, wang2019evidence}, $o_s$ \emph{is not explicitly trained for this purpose}.
Concretely, our analysis covers three settings, illustrated in Figure~\ref{figure:overview} and summarized in Table~\ref{table:settings}.

The first setting (Figure~\ref{figure:overview} left, and \S\ref{sec:numerical_arguments_extraction}) considers a model with generative and extractive heads, trained on the \drop{} dataset \cite{dua2019drop} for numerical reasoning over text. Surprisingly, we observe that the arguments for the arithmetic computation required for the generative head to generate its answer often emerge in the outputs of the extractive head. The second setting (Figure~\ref{figure:overview} middle, and \S\ref{sec:supporting_facts_extraction}) considers a model with a classification head outputting \emph{`yes'}/\emph{`no'} answers, and a span extraction head, trained on the \hotpotqa{} dataset \cite{yang2018hotpotqa} for multi-hop reasoning. The outputs of the extractive head once again provide explanations in the form of supporting facts from the input context. 
The last setting (Figure~\ref{figure:overview} right, and \S\ref{sec:query_based_summarization}) considers a model with two extractive heads, one for span extraction and another for (sentence-level) extractive summarization. Each head is trained on a different dataset; \hotpotqa{} for span extraction and \cnndaily{} for summarization \cite{hermann2015teaching}. We find that the summarization head tends to extract the supporting facts given inputs from \hotpotqa{}, effectively acting as a query-based summarization model.

\commentout{
The first setting (Fig.~\ref{figure:overview}, left) considers a model with generative and extractive heads, trained on the \drop{} dataset for numerical reasoning over text \cite{dua2019drop} (\S\ref{sec:numerical_arguments_extraction}). Surprisingly, we observe that the arguments for the arithmetic computation required for the generative head to generate its answer often emerge in the outputs of the extractive head. We measure the coverage of arguments and show that successful argument extraction is correlated with performance. Moreover, we show that by examining the output of the extractive head, we can intervene and modify the behaviour of the target head in a predictable manner. Last, we show that this behaviour is tied to the size of the two heads, and generalizes to out-of-distribution samples \jb{don't know what you mean here}. 

The second setting (Fig.~\ref{figure:overview}, middle) considers a model with classification and extractive heads, trained on the \hotpotqa{} dataset \cite{yang2018hotpotqa} for multi-hop reasoning (\S\ref{sec:supporting_facts_extraction}). Examining the predictions of the extractive head, reveals once again that the it provides explanations for the model predictions, this time in the form of supporting facts from the input context. 

The last setting (Fig.~\ref{figure:overview}, right) considers a model with two extractive heads, one for span extraction and another one for extractive summarization at the sentence level (\S\ref{sec:query_based_summarization}). Each head is trained on a different dataset, \hotpotqa{} for span extraction and \cnndaily{} \cite{see2017point} for summarization. We find that despite the different distributions, the summarization head tends to extract the supporting facts given inputs from \hotpotqa{}, effectively acting as a query-based summarizer.
}

We now present the above settings. Table~\ref{table:settings} summarizes the main results. We denote by $\textit{FFNN}^{(l)}_{m\times n}$ a feed-forward neural network with $l$ layers that maps inputs of dimension $m$ to dimension $n$.


\begin{table*}[t]
    \centering
    \footnotesize
    \begin{tabular}{l|c|c|c|c|c||c|c}
          & \% Qs with & Recall & Precision & Correlation & Avg. \# & \drop{} & \drop{}$_{\text{span}}$  \\
          & recall=1.0 & & & & of spans & F$_1$ & F$_1$  \\\hline
         \msegenbert{} & \textbf{0.41} & \textbf{0.56} & 0.6 & \textbf{0.35} & 2.1 & 70.4 & 63.0  \\
         \msebert{} & 0.1 & 0.2 & 0.32 & 0.17 & 1.0 & 35.6 & 64.3 \\
         \hline
         \msegenbert{}$_{l=2}$ & 0.26 & 0.48 & \textbf{0.72} & 0.32 & 1.2 & 70.4 & 64.1 \\
         \msegenbert{}$_{l=4}$ & 0.24 & 0.47 & 0.71 & 0.33 & 1.2 & 70.6 & 63.7 \\
         \msegenbert{}$_{\text{decoder untied}}$ & 0.27 & 0.48 & 0.69 & \textbf{0.35} & 1.2 & 71.0 & 64.8
    \end{tabular}
    \caption{Evaluation results of \msegenbert{}. \drop{} F$_1$ and \drop{}$_{\text{span}}$ F$_1$ were computed on the \drop{} development set and its subset of examples with span answers, respectively. All other scores are on the 400 annotated examples from \drop{}. Correlation is between recall and F$_1$ scores. $l$ refers to the number of linear layers in $o_{\textit{mse}}$.}
    \label{table:genbert_results}
\end{table*}

\begin{table}[t]
    \centering
    \footnotesize
    \begin{tabular}{p{7.3cm}}
          \hline
          \textbf{Passage}: According to the 2014 census, 1,144,428 residents or 38,2\% live in cities while 1,853,807 are rural residents. The largest cities under the control of the constitutional authorities are Chisinau with 644,204 (with 590,631 actual urban dwellers) and Balti with 102,457 (97,930 urban dwellers). The autonomous territorial unit of Gagauzia has 134,535, out of which 48,666 or 36,2\% are urban dwellers. Ungheni is the third largest city with 32,828, followed by Cahul with \textbf{28,763}, Soroca with \textbf{22,196} and Orhei with \textbf{21,065}. \\
          \textbf{Question}: How many people are in Cahul,
          Soroca, and Orhei combined? (72,024) \\
          \textbf{Arguments}: 28,763, 22,196, 21,065 \\ \hline
    \end{tabular}
    \caption{Annotated example from \drop{}. Crowdworkers were asked to extract the arguments (in bold) from the passage required to compute the answer.}
    \label{table:example_drop_annotation}
\end{table}


\section{Setting 1: Emerging Computation Arguments in Span Extraction}
\label{sec:numerical_arguments_extraction}

We start by examining a combination of generative and extractive heads (Figure~\ref{figure:overview}, left), and analyze the spans extracted from the input when the generative head is selected to output the final answer.

\subsection{Experimental Setting}
\paragraph{Model}
We take \genbert{} \cite{geva2020injecting}, a BERT-base model fine-tuned for numerical reasoning, and use it to initialize a variant called \msegenbert{}, in which the single-span extraction head is replaced by a multi-span extraction (MSE) head introduced by \citet{segal2020simple}, which allows extracting multiple spans from the input. This is important for supporting extraction of more than one argument. \msegenbert{} has three output heads: The multi-span head, which takes $\HH \in \mathbb{R}^{d \times n}$, and uses the \texttt{BIO} scheme \cite{ramshaw1995text} to classify each token in the input as the beginning of (\texttt{B}), inside of (\texttt{I}), or outside of (\texttt{O}) an answer span:
$$ o_{\textit{mse}} := \textit{FFNN}^{(1)}_{d\times 3}(\HH) \in \mathbb{R}^{3 \times n}. $$
The second head is the generative head, \ogen{}, a standard transformer decoder \cite{vaswani2017attention} initialized by BERT-base, that is tied to the encoder and performs cross-attention over $\HH$ \cite{geva2020injecting}.
Last, a classification head takes the representation $\hh_\text{CLS}$ of the \texttt{CLS} token and  selects the target head ($o_\textit{mse}$ or $o_\textit{gen}$):
$$ o_{\textit{type}} := \textit{FFNN}^{(1)}_{d\times 2}(\hh_{\text{CLS}}) \in \mathbb{R}^2. $$
Implementation details are in Appendix~\ref{subsec:fine_tuning_details_drop}.

\paragraph{Data}
We fine-tune \msegenbert{} on \drop{} \cite{dua2019drop}, a dataset for numerical reasoning over paragraphs, consisting of passage-question-answer triplets where answers are either spans from the input or numbers that are not in the input.
Importantly, \omse{} is trained only on span questions, as its outputs are restricted to tokens from the input. Moreover, less than 5\% of \drop{} examples have multiple spans as an answer.

To evaluate the outputs of \omse{} on questions where the answer is a number that is not in the input, we use crowdsourcing to annotate 400 such examples from the development set. Each example was annotated with the arguments to the computation that are in the passage and are required to compute the answer. 
Each example was annotated by one of 7 crowdworkers that were qualified for the task. We regularly reviewed annotations to give feedback and avoid bad annotations.
An example annotation is provided in Table~\ref{table:example_drop_annotation}. On average, there are 1.95 arguments annotated per question.

\paragraph{Evaluation metrics}
Given a list $\cP$ of extracted spans by \omse{} and a list of annotated arguments $\cG$, we define the following metrics for evaluation:\footnote{For arguments and spans which include a number as well as text, only the numeric sub-strings were considered when preforming the comparison between $\cP$ and $\cG$.} We check argument recall by computing the fraction of arguments in $\cG$ that are also in $\cP$: $\frac{|\cP\ \cap\ \cG|}{|\cG|}$.
We can then compute average recall over the dataset, and the proportion of questions with a perfect recall of 1.0 (first column in Table~\ref{table:genbert_results}). 
Similarly, we compute precision by computing the fraction of arguments in $\cP$ that are also in $\cG$: $\frac{|\cP\ \cap\  \cG|}{|\cP|}$ and then the average precision over the dataset. 

\subsection{Results}
Table~\ref{table:genbert_results} presents the results. Comparing \msegenbert{} to \msebert{}, where the model was trained without \ogen{} only on span extraction examples, we observe that multi-task training substantially changes the behaviour of the extractive head. First, \msegenbert{} dramatically improves the extraction of computation arguments: recall increases from 0.2$\rightarrow$0.56, precision goes up from 0.32$\rightarrow$0.6, and the fraction of questions with perfect recall reaches 0.41. The number of extracted spans also goes up to 2.1, despite the fact that most span extraction questions are a single span.
The performance of \msebert{} on span questions is similar to \msegenbert{}, showing that the difference is not explained by performance degradation.
We also measure the number of extracted spans on out-of-distribution math word problems, and observe similar patterns (details are in Appendix~\ref{subsec:num_spans_mwp}).

Moreover, model performance, which depends on \ogen{}, is correlated with the recall of predicted spans, extracted by \omse{}. The Spearman correlation between model F$_1$ and recall for \msegenbert{} is high at 0.351 (Table~\ref{table:genbert_results}) and statistically significant (p-value $5.6e^{-13}$), showing that when the computation arguments are covered, performance is higher.

These findings illustrate that multi-task training leads to emergent behaviour in the extractive head, which outputs computation arguments for the output of the generative head. We now provide more fine-grained analysis.

\paragraph{Distribution of extracted spans}
On average, \msegenbert{} extracts 2.12 spans per example, which is similar to 1.95 spans extracted by annotators. 
Moreover, the average ratio $\frac{|\cP|}{|\cG|}$ is 1.2, indicating good correlation at the single-example level.
Table~\ref{table:examples_args_drop} shows example outputs of \omse{} vs. the annotated arguments for the same questions. 
The full distributions of the number of extracted spans by \msegenbert{} compared to the annotated spans are provided in Appendix~\ref{subsec:num_spans_drop}.

\begin{table}[t]
     \centering
     \footnotesize
     \begin{tabular}{l|c|c}
         & Annotated & Predicted \\
         & arguments & arguments \\ \hline
         Full match  & 26, 48 & 26, 48 \\
         \omse{} missing & 31.7, 20.1 & 31.7 \\
         \omse{} excessive & 1923, 1922 & 1923, 1937, 1922 \\
     \end{tabular}
     \caption{Example outputs by \omse{} on \drop{} in comparison to the annotated computation arguments.}
     \label{table:examples_args_drop}
 \end{table}

\paragraph{Parameter sharing across heads}
We conjecture that the steering effect occurs when the heads are strongly tied, with most of their parameters shared. To examine this, we increase the capacity of the FFNN in $o_\textit{mse}$ from $l=1$ layer to $l=2,4$ layers, and also experiment with a decoder whose parameters, unlike \genbert{}, are not tied to the encoder.

We find (Table~\ref{table:genbert_results}) that reducing the dependence between the heads also diminishes the steering effect. While the models still tend to extract computation arguments, with much higher recall and precision compared to \msebert{}, they output 1.2 spans on average, which is similar to the distribution they were trained on. This leads to higher precision, but much lower recall and fewer cases of prefect recall. Overall model performance is not affected by changing the capacity of the heads.

\begin{figure}[t]
    \centering
    \includegraphics[scale=0.43]{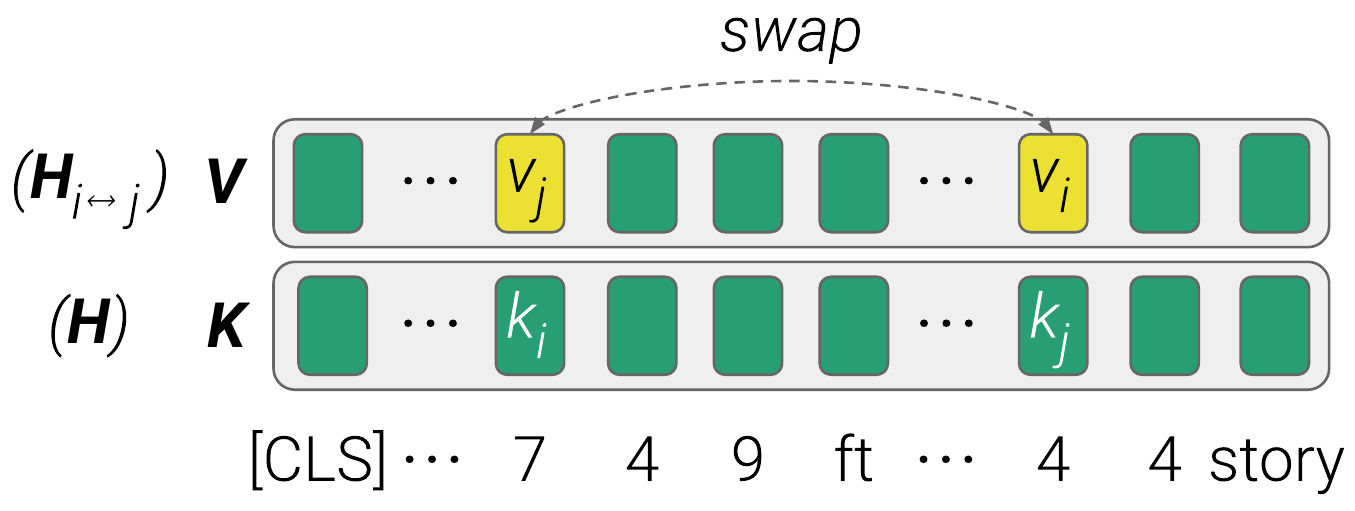}
    \caption{Illustration of our intervention method, where two value vectors $\vv_i,\vv_j$ are swapped in cross-attention.}
    \label{figure:hs_swap}
\end{figure}

\subsection{Influence of Extracted Spans on Generation}
The outputs of \omse{} and \ogen{} are correlated, but can we somehow \emph{control} the output of \ogen{} by modifying the value of span tokens extracted by \omse{}?
To perform such intervention, we change the cross-attention mechanism in \msegenbert{}'s decoder. Typically, the keys and values are both the encoder representations $\HH$. To modify the values read by the decoder, we change the value matrix to $\HH_{i \leftrightarrow j}$:  
$$ \textit{MultiHeadAttention}(\mathbf{Q}, \HH, \HH_{i \leftrightarrow j}), $$ where in $\HH_{i \leftrightarrow j}$ the  positions of the representations $\hh_i$ and $\hh_j$ are swapped (illustrated in Figure~\ref{figure:hs_swap}). Thus, when the decoder attends to the $i$'th token, it will get the value of the $j$'th token and vice versa.

We choose the tokens $i,j$ to swap based on the output of \omse{}. Specifically, for every input token $x_k$ that is a digit,\footnote{\genbert{} uses digit tokenization for numbers.} we compute the probability $p_k^{\texttt{B}}$ by \omse{} that it is a beginning of an output span. Then, we choose the position $i = \arg\max_k p_k^{\texttt{B}}$, and the position $j$ as a random position of a digit token. 
As a baseline, we employ the same procedure, 
but swap the positions $i,j$ of the two digit tokens with the highest outside (\texttt{O}) probabilities.

We focus on questions where \msegenbert{} predicted a numeric output. Table~\ref{table:swap_change_unchange} shows in how many cases each intervention instance changed the model prediction (by \ogen{}). For 40.6\% of the questions, the prediction was altered due to the intervention in the highest probability \texttt{B} token, compared to only 0.03\% (2 cases) by the baseline intervention. This shows that selecting the token based on \omse{} affects whether this token will lead to a change.

More interestingly, we test whether we can predict the change in the output of \ogen{} by looking at the
two digits that were swapped. 
Let $d$ and $d'$ be the values of digits swapped, and let $n$ and $n'$ be the numeric outputs generated by \ogen{} before and after the swap. We check whether $|n- n'| = |d - d'|*10^c$ for some integer $c$. For example, if we swap the digits `7' and `9', we expect the output to change by 2, 20, 0.2, etc. We find that in 543 cases out of 2,460 (22.1\%) the change in the model output is indeed predictable in the non-baseline intervention, which is much higher than random guessing, that would yield 10\%.

Last, we compare the model accuracy on predictable and unpredictable cases, when intervention is not applied to the examples. We observe that exact-match performance is 76\% when the change is predictable, but only 69\% when it is not. This suggests that interventions lead to predictable changes with higher probability when the model is correct. 

Overall, our findings show that the spans extracted by \omse{} affect the output of \ogen{}, while spans \omse{} marks as irrelevant do not affect the output. Moreover, the (relative) predictability of the output after swapping shows that the model performs the same computation, but with a different argument.


\begin{table}[t]
    \centering
    \footnotesize
    \begin{tabular}{l|cc|c}
         & \texttt{B}-swap &  \texttt{B}-swap \\
         & changed &  unchanged \\\hline
         \texttt{O}-swap changed & 0.03 & 0.0  & 0.03 \\ 
         \texttt{O}-swap unchanged & 40.62 & 59.35 & 99.97
 \\ \hline
         & 40.65 & 59.35 & 100
    \end{tabular}
    \caption{Intervention results on \ogen{} outputs. Percentage out of 6,051 \drop{}'s development examples for which swapping \texttt{B} (or \texttt{O}) tokens changed (or did not change) the output of \msegenbert{}.}
    \label{table:swap_change_unchange}
\end{table}


\section{Setting 2: Emerging Supporting Facts in Span Extraction}
\label{sec:supporting_facts_extraction}

We now consider a combination of an extractive head and a classification head (Figure~\ref{figure:overview}, middle).

\subsection{Experimental Setting}
\paragraph{Model}
We use the BERT-base \reader{} model introduced by \citet{asai2020Learning}, which has two output heads: A single-span extraction head, which predicts for each token the probabilities for being the \texttt{start} and \texttt{end} position of the answer span:
$$ o_{\textit{sse}} := \textit{FFNN}^{(1)}_{d\times 2}(\HH). $$
The second head is a classification head for the answer type: \texttt{yes}, \texttt{no}, \texttt{span}, or \texttt{no-answer}:
$$ o_{\textit{type}} := \textit{FFNN}^{(1)}_{d\times 4}(\hh_{\text{CLS}}). $$
Implementation details are in Appendix~\ref{subsec:fine_tuning_details_for_hotpot_alone}.

\paragraph{Data}
We fine-tune a \reader{} model on the gold paragraphs of \hotpotqa{} \cite{yang2018hotpotqa}, a dataset for multi-hop QA. Specifically, we feed the model question-context pairs and let it predict the answer type with \otype{}. If \otype{} predicts \texttt{span} then the output by \osse{} is taken as the final prediction, otherwise it is the output by \otype{}. Therefore, \osse{} is trained only on examples with an answer span.

Examples in \hotpotqa{} are annotated with \emph{supporting facts}, which are sentences from the context that provide evidence for the final answer. We use the supporting facts to evaluate the outputs of \osse{} as explanations for questions where the gold answer is \texttt{yes} or \texttt{no}.

\begin{table}[t]
    \centering
    \footnotesize
    \begin{tabular}{l|c|c|c}
          & \% Qs with & Inverse  & \hotpotqa{} \\
          & Recall@5=1 & MRR  & F$_1$ \\ \hline
         \reader{}  & \textbf{0.605} & \textbf{0.867}  & 72.9 \\
         \reader{}$_{\text{only sse}}$ & 0.539 & 0.828  & 70.5 \\ \hline
         \reader{}$_{l=2}$  & 0.568 & 0.860  & 73.7
         
    \end{tabular}
    \caption{Evaluation results of \reader{}. F$_1$ scores were computed over the development set of \hotpotqa{}, and the rest of the scores on the development subset of yes/no questions, using $k=5$. The parameter $l$ refers to the number of linear layers in each of \osse{} and \otype{}.}
    \label{table:reader_results}
\end{table}

\paragraph{Evaluation metrics}
Let $\cF$ be the set of annotated supporting facts per question and $\cP$ be the top-$k$ output spans of \osse{}, ordered by decreasing probability.
We define Recall@k to be the proportion of supporting facts covered by the top-k predicted spans, where a fact is considered covered if a predicted span is within the supporting fact sentence and is not a single stop word (see Table~\ref{table:example_sf}).\footnote{We do not define coverage as the fraction of tokens in a supporting fact that the span covers, because supporting facts are at the sentence-level, and often times most of the tokens in the supporting fact are irrelevant for the answer.}
We use $k=5$ and report the fraction of questions where Recall@5 is 1 (Table~\ref{table:reader_results}, first column), to measure the cases where \osse{} covers \emph{all} relevant sentences in the first few predicted spans.

Additionally, we introduce an InverseMRR metric, based on the MRR measure, as a proxy for precision. We take the rank $r$ of the first predicted span in $\cP$ that is not a supporting fact from $\cF$, and use $1 - \frac{1}{r}$ as the measure (e.g., if the rank of the first non overlapping span is 3, the reciprocal is 1/3 and the InverseMRR is 2/3).
If the first predicted span is not in a supporting fact, InverseMRR is 0; if all spans for $k=5$ overlap, InverseMRR is 1.


\begin{table}[t]
    \centering
    \footnotesize
    \begin{tabular}{p{7.3cm}}
          \toprule
          \textbf{Question}: Were Goo Goo Dolls and Echosmith formed in the same city? (\nl{no}) \\
          \hline
          \textbf{1.} Goo Goo Dolls \\
          \textbf{2.} Goo Goo Dolls are an American rock band formed in 1985 in Buffalo, New York \\
          \textbf{3.} Echosmith is an American, Corporate indie pop band formed in February 2009 in Chino, California \\
          \textbf{4.} New York \\
          \textbf{5.} Chino, California
          \\ \toprule
          \toprule
          \textbf{Question}: Is the building located at 200 West Street taller than the one at 888 7th Avenue? (\nl{yes}) \\
          \hline
          \textbf{1.} building \\
          \textbf{2.} building is a 749 ft , 44-story building \\
          \textbf{3.} The building \\
          \textbf{4.} The building is a 749 ft , 44-story building \\
          \textbf{5.} 888 7th Avenue is a 628 ft (191m) tall
          \\ \toprule
    \end{tabular}
    \caption{Example questions from \hotpotqa{} and the top-5 spans extracted by the \reader{} model.}
    \label{table:example_sf}
\end{table}

\subsection{Results}
Results are presented in Table~\ref{table:reader_results}.
Comparing \reader{} and \reader{}$_{\text{only sse}}$, the Recall@5 and InverseMRR scores are substantially higher when using multi-task training, with an increase of 10.9\% and 4.5\%, respectively, showing again that multi-task training is the key factor for emerging explanations.
Example questions with the spans extracted by \reader{} are provided in Table~\ref{table:example_sf}.

As in \S\ref{sec:numerical_arguments_extraction},
adding an additional layer to \osse{} (\reader{}$_{l=2}$) decreases  the frequency of questions with perfect Recall@5 (0.605 $\rightarrow$ 0.568). This shows again that reducing the dependency between the heads also reduces the steering effect.
It is notable that the performance on \hotpotqa{} is similar across the different models, with only a slight deterioration when training only the extraction head (\osse{}). This is expected as \reader{}$_{\text{only sse}}$ is not trained with yes/no questions, which make up a small fraction of \hotpotqa{}.




\section{Setting 3: Emerging Query-based Summaries}
\label{sec:query_based_summarization}

In \S\ref{sec:numerical_arguments_extraction} and \S\ref{sec:supporting_facts_extraction}, we considered models with output heads trained on examples from the same data distribution. 
Would the steering effect occur when output heads are trained on different datasets? We now consider a model trained to summarize text and answer multi-hop questions (Figure~\ref{figure:overview}, right).

\subsection{Experimental Setting}

\paragraph{Model}
We create a model called \readersum{} as follows: We take the \textsc{Reader} model from \S\ref{sec:supporting_facts_extraction}, and add the classification head presented by \citet{liu2019text}, that summarizes an input text by selecting sentences from it. Sentence selection is done by inserting a special \texttt{[CLS]} token before each sentence and training a summarization head to predict a score for each such \texttt{[CLS]} token from the representation $\hh_{CLS}$:
$$ o_{\textit{sum}} := \textit{FFNN}_{d\times 1}^{(1)}(\hh_{\text{CLS}}). $$
The sentences are ranked by their scores and the top-3 highest score sentences are taken as the summary (top-3 because choosing the first 3 sentences of a document is a standard baseline in extractive summarization
 \cite{nallapati2017summarunner, liu2019text}).
Implementation details are in~\ref{subsec:fine_tuning_details_for_hotpot_summary}.

\paragraph{Data}
The QA heads (\osse{}, \otype{}) are trained on \hotpotqa{}, while the summarization head is trained on the \cnndaily{} dataset for extractive summarization \cite{hermann2015teaching}. We use the supporting facts from \hotpotqa{} to evaluate the outputs of \osum{} as explanations for predictions of the QA heads.

\paragraph{Evaluation metrics}
Annotated supporting facts and the summary are defined by sentences from the input context. Therefore, given a set $\cT$ of sentences extracted by \osum{} ($|\cT|=3$) and the set of supporting facts $\cF$, we compute the Recall@3 of $\cT$ against $\cF$. 

\begin{table}[t]
    \centering
    \footnotesize
    \begin{tabular}{l|c||c|c}
          & Recall@3 &  F$_1$ & ROUGE 1/2  \\ \hline
         \readersum{} & \textbf{0.79} & 71.6 & 42.7/19.2 \\
         \readersum{}$_{\text{only sum}}$ & 0.69 & - & 43.6/19.9 \\ \hline
         \textsc{Random} & 0.53 & - & 32.1/10.9 \\
         \textsc{Lead3} & 0.60 & - & 40.6/17.1 \\
         \readersum{}$_{\text{masked}}$ & 0.66 & - & 42.7/19.2
    \end{tabular}
    \caption{Evaluation results of \readersum{}. Recall@3 and F$_1$ scores were computed over the development set of \hotpotqa{}, and ROUGE over \cnndaily{}.}
    \label{table:reader_summarizer_results}
\end{table}

\subsection{Results}

Results are summarized in Table~\ref{table:reader_summarizer_results}.
When given \hotpotqa{} examples, \readersum{}  extracts summaries that cover a large fraction of the supporting facts (0.79 Recall@3). 
This is much higher compared to a model that is trained only on the extractive summarization task (\readersum{}$_{\text{only sum}}$ with 0.69 Recall@3). 
Results on \cnndaily{} show that this behaviour in \readersum{} does not stem from an overall improvement in extractive summarization, as \readersum{} performance is slightly lower compared to \readersum{}$_{\text{only sum}}$.

To validate this against other baselines, both \readersum{} and \readersum{}$_{\text{only sum}}$ achieved substantially better Recall@3 scores compared to a baseline that extracts three random sentences from the context (\textsc{Random} with $0.53$ Recall@3), and summaries generated by taking the first three sentences of the context (\textsc{Lead3} with $0.6$ Recall@3).

Overall, the results show multi-head training endows \osum{} with an emergent behavior of query-based summarization, which we evaluate next.
Example summaries extracted by \readersum{} for \hotpotqa{} are provided in Appendix~\ref{sec:example_emergent_summaries}.

\paragraph{Influence of questions on predicted summaries}
We run \readersum{} on examples from \hotpotqa{} while masking out the questions, thus, \osum{} observes only the context sentences.
As shown in Table~\ref{table:reader_summarizer_results} (\readersum{}$_{\text{masked}}$), masking the question leads to a substantial decrease of 13 Recall@3 points in comparison to the same model without masking (0.79$\rightarrow$0.66).  

\begin{figure}[t]
    \centering
    \includegraphics[scale=0.39]{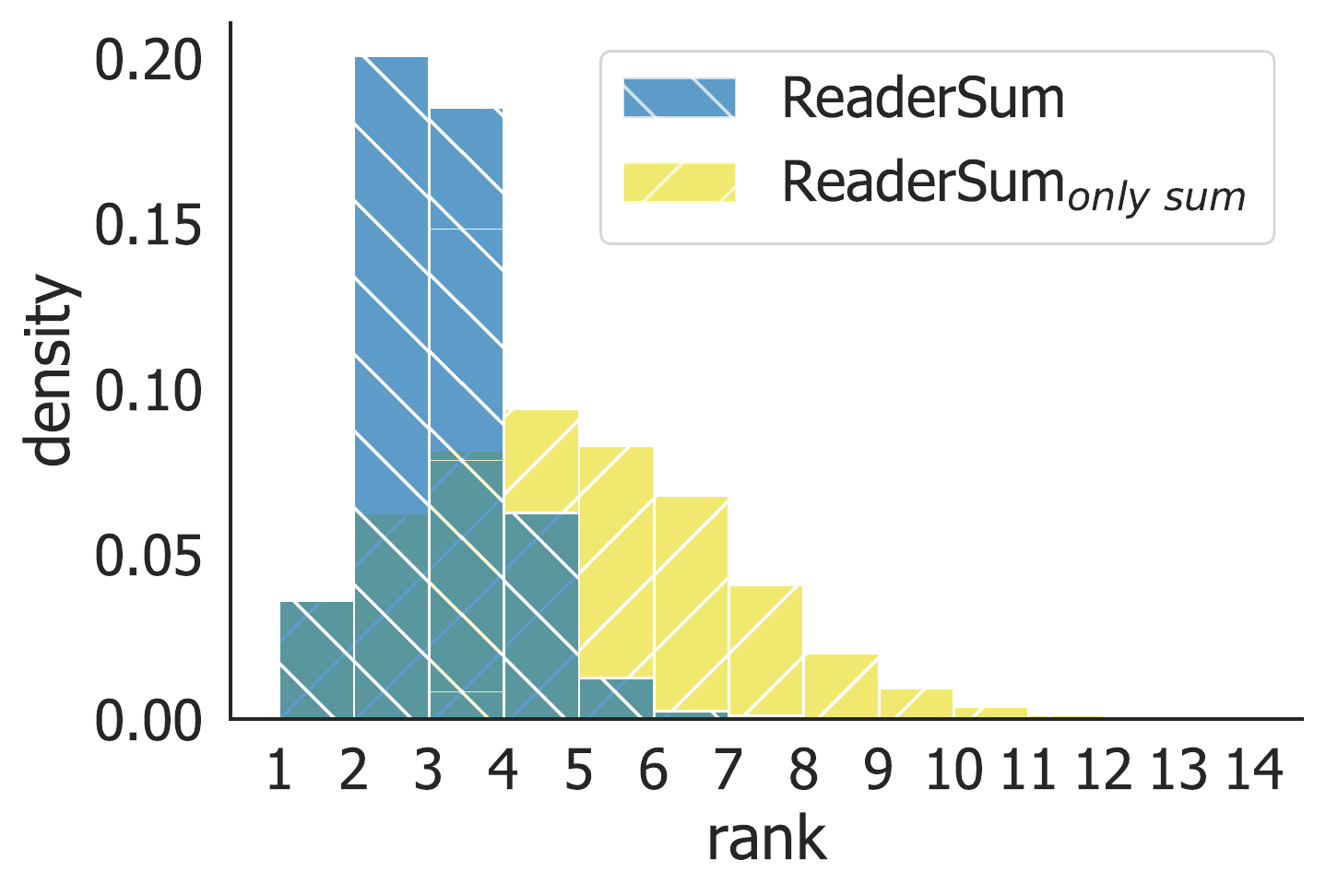}
    \caption{Distribution over ranks for \hotpotqa{} question sentences, based on scores predicted by \osum{}, for \readersum{} and \readersum{}$_{\text{only sum}}$.}
    \label{figure:score_dist}
\end{figure}

Since our model appends a \texttt{[CLS]} token to every sentence, including the question (which never appears in the summary), we can rank the question sentence based on the score of \osum{}. Computing the rank distribution of question sentences, we see (Figure~\ref{figure:score_dist}) that the distributions of \readersum{}$_{\text{only sum}}$ and \readersum{} are significantly different,\footnote{Wilcoxon signed-rank test\commentout{statistic $1908795$} with p-value $\ll0.001$.} and that questions are ranked higher in  \readersum{}. This shows that the summarization head puts higher emphasis on the question in the multi-head setup.

Overall, these results provide evidence that multi-head training pushes \osum{} to perform query-based summarization on inputs from \hotpotqa{}.

\section{Related Work}
\label{sec:related_work}

Transformer models with multiple output heads have been widely employed in previous works \cite{hu2021transformer, aghajanyan2021muppet, segal2020simple, hu2019multi, clark2019bam}. To the best of our knowledge, this is the first work that analyzes the outputs of the non-target heads.

Previous work used additional output heads to generate explanations for model predictions \cite{perez2019finding, schuff2020f1, wang2019evidence}. Specifically, recent work has explored utilization of summarization modules for explainable QA \cite{nishida2019answering, deng2020joint}.
In the context of summarization, \citet{xu2020coarse} have leveraged QA resources for training query-based summarization models.
Hierarchies between NLP tasks have also been explored in multi-task models not based on transformers \cite{sogaard2016deep, hashimoto2017joint, swayamdipta2018syntactic}.
Contrary to previous work, the models in this work were \emph{not trained} to perform the desired behaviour. Instead, explanations and generalized behaviour \emph{emerged} from training on multiple tasks. 

A related line of research has focused on developing probes, which are supervised network modules that predict properties from model representations \cite{conneau2018cram, van2019how,tenney2019bert,liu2019linguistic}. A key challenge with probes is determining whether the information exists in the representation or is learned during probing \cite{hewitt2019designing, tamkin2020investigating,talmor2020olmpics}.
Unlike probes, steered heads are trained in parallel to target heads rather than on a fixed model. 
Moreover, steered heads are not designed to decode specific properties from representations, but their behaviour naturally extends beyond their training objective.



Our findings also relate to explainability methods that highlight parts from the input via the model's attention \cite{wiegreffe2019attention}, and extract rationales through unsupervised training \cite{lei2016rationalizing}. The emerging explanations we observe are based on the predictions of a head rather than on internal representations.

\section{Conclusions and Discussion}
\label{sec:conclusion}
We show that training multiple heads on top of a pre-trained language model creates a steering effect, where the target head influences the behaviour of another head, steering it towards capabilities beyond its training objective.
In three multi-task settings, we find that without any dedicated training, the steered head often outputs explanations for the model predictions. Moreover, modifying the input representation based on the outputs of the steered head can lead to predictable changes in the target head predictions. 

Our findings provide evidence for extrapolation of skills as a consequence of multi-task training, opening the door to new research directions in interpretability and generalization.
Future work could explore additional head combinations, in order to teach models new skills that can be cast as an extrapolation of existing tasks. In addition, the information decoding behaviour observed in this work can serve as basis for developing general interpretability methods for debugging model predictions. 

A natural question that arises is what head combinations lead to a meaningful steering effect.
We argue that there are two considerations involved in answering this question. First, the relation between the \emph{tasks} the heads are trained on. The tasks should complement each other (e.g. summarization and question answering), or the outputs of one task should be expressive enough to explain the outputs of the other task, when applied to inputs of the other task.
For example, extractive heads are particularly useful when the model's output is a function of multiple input spans.
Another consideration is the \emph{inputs} to the heads. We expect that training heads with similar inputs (in terms of length, language-style, etc.) will make the underlying language model construct similar representations, thus, increasing the probability of a steering effect between the heads.

\section*{Acknowledgements}
We thank Ana Marasovi\'c and Daniel Khashabi for the helpful feedback and constructive suggestions, and the NLP group at Tel Aviv University, particularly Maor Ivgi and Elad Segal. This research was supported in part by The Yandex Initiative for Machine Learning, and The European Research Council (ERC) under the European Union Horizons 2020 research and innovation programme (grant ERC DELPHI 802800). This work was completed in partial fulfillment for the Ph.D degree of Mor Geva.

\bibliography{all}
\bibliographystyle{acl_natbib}


\appendix


\section{Implementation Details}
\label{sec:fine_tuning_details}

\subsection{Models Trained on \drop{}}
\label{subsec:fine_tuning_details_drop}
We implement \msegenbert{} by taking \genbert{}\footnote{\url{https://github.com/ag1988/injecting_numeracy/}} \cite{geva2020injecting} and replacing its span-extraction head with the tagging-based multi-span extraction head\footnote{\url{https://github.com/eladsegal/tag-based-multi-span-extraction}} by \citet{segal2020simple}.

All the variants of \msegenbert{} were initialized with the checkpoint of \genbert{}$_{\textsc{+ND+TD}}$, that was fine-tuned on both numerical and textual data. For fine-tuning on \drop{}, we used the same hyperparameters used in  \citet{geva2020injecting}, specifically, a learning rate of $3e^{-5}$ with linear warmup $0.1$ and weight decay $0.01$ for $30$ epochs, and a batch size of $16$.

\subsection{Models Trained on \hotpotqa{} Alone}
\label{subsec:fine_tuning_details_for_hotpot_alone}
To train \reader{} models on \hotpotqa{}, we used the official code\footnote{\label{footnote:reader_code}\url{https://github.com/AkariAsai/learning_to_retrieve_reasoning_paths}} by \citet{asai2020Learning}. We fine-tuned the models on the gold paragraphs (without the distractor paragraphs) for 2 epochs with a learning rate of $5e^{-5}$ and a batch size $16$. All the other hyperparameters remained the same as in the implementation of \citet{asai2020Learning}.

\subsection{Models Trained on \hotpotqa{} and \cnndaily{}}
\label{subsec:fine_tuning_details_for_hotpot_summary}
To train a model for both question answering and summarization, we used the official code\textsuperscript{\ref{footnote:reader_code}} by \citet{asai2020Learning}, but adapted it to serve also for the extractive summarization task, by adding the classification head of \textsc{BERTSum} \footnote{\url{https://github.com/nlpyang/BertSum}} \cite{liu2019text}. 
\textsc{BERTSum} summarizes an input context by selecting sentences (see \S\ref{sec:query_based_summarization}). To allow this mechanism, we modify the inputs from \hotpotqa{} and \cnndaily{} as follows: First, we split the context into sentences using Stanza \cite{qi2020stanza}. Then, a special \texttt{[CLS]} token is added at the beginning of each sentence and a \texttt{[SEP]} token is added at the end of it. 

The model is fine-tuned for each task by training on one batch from each task at a time, i.e. a batch of \hotpotqa{} followed by a batch of \cnndaily{}. To obtain the oracle summaries for \cnndaily{}, we used the greedy algorithm described in \cite{nallapati2016abstractive}. The model trained with a learning rate of $5e^{-5}$ and batch size 8 for 1 epoch.

\section{Distribution of Extracted Spans by \msegenbert{}}
\label{sec:num_span_mse}

\subsection{Distribution of Extracted and Annotated Spans from \drop{}}
\label{subsec:num_spans_drop}
\begin{figure}[t]
    \centering
    \includegraphics[scale=0.39]{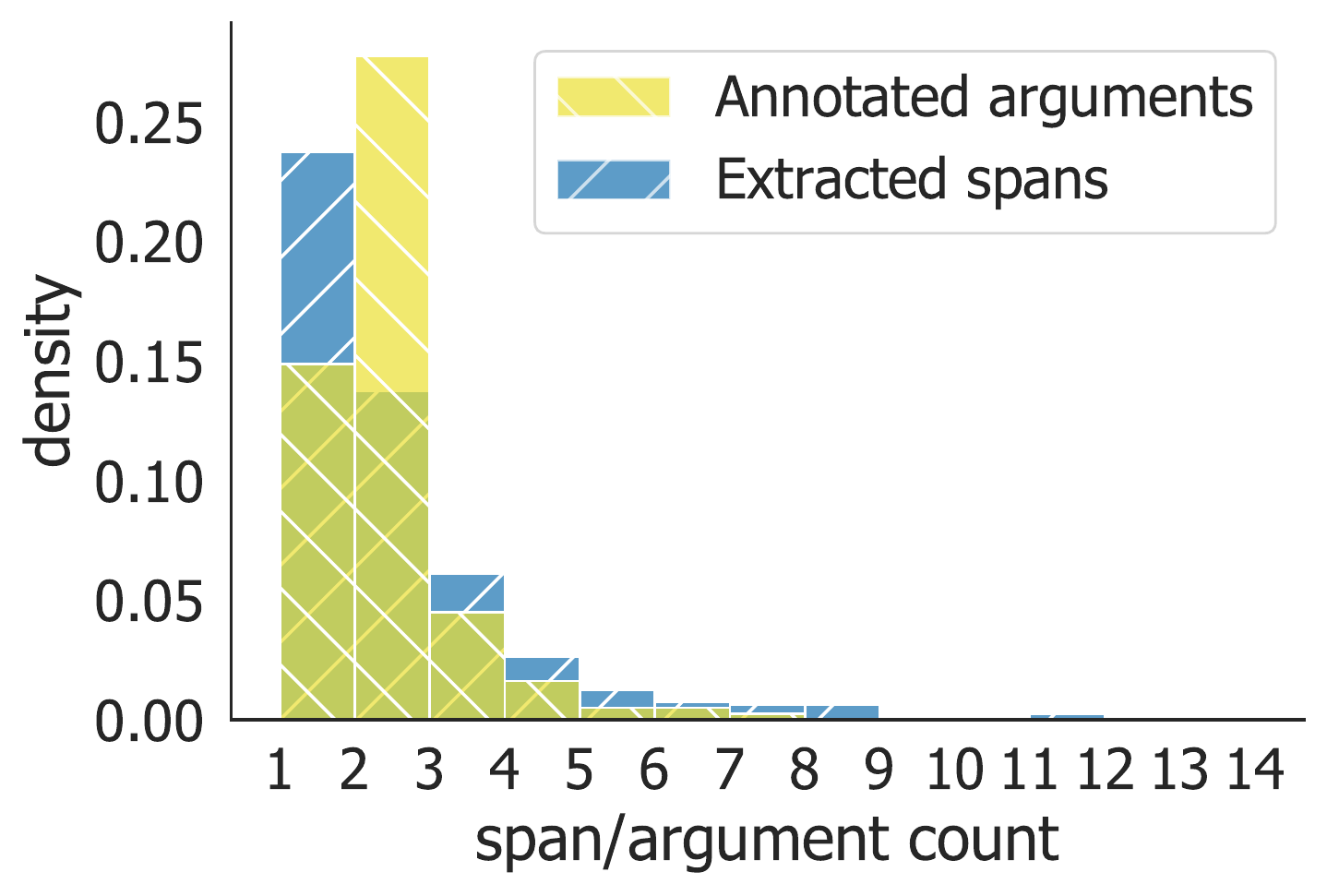}
    \caption{The number of spans extracted by \msegenbert{} \omse{} vs. the number of annotated arguments for the same questions.}
    \label{figure:num_spans_annotated_extracted}
\end{figure}

Figure~\ref{figure:num_spans_annotated_extracted} shows the number of extracted spans by \msegenbert{} compared to the annotated spans, for a subset of 400 examples from the development set of \drop{}.
On average, \msegenbert{} extracts a similar number of spans per example (2.12) compared to the spans extracted by annotators (1.95). However, \msegenbert{} tends to over-predict one span compared to the annotated examples.

\subsection{Distribution of Extracted Spans on Math-Word-Problems}
\label{subsec:num_spans_mwp}
\begin{figure}[t]
    \centering
    \includegraphics[scale=0.4]{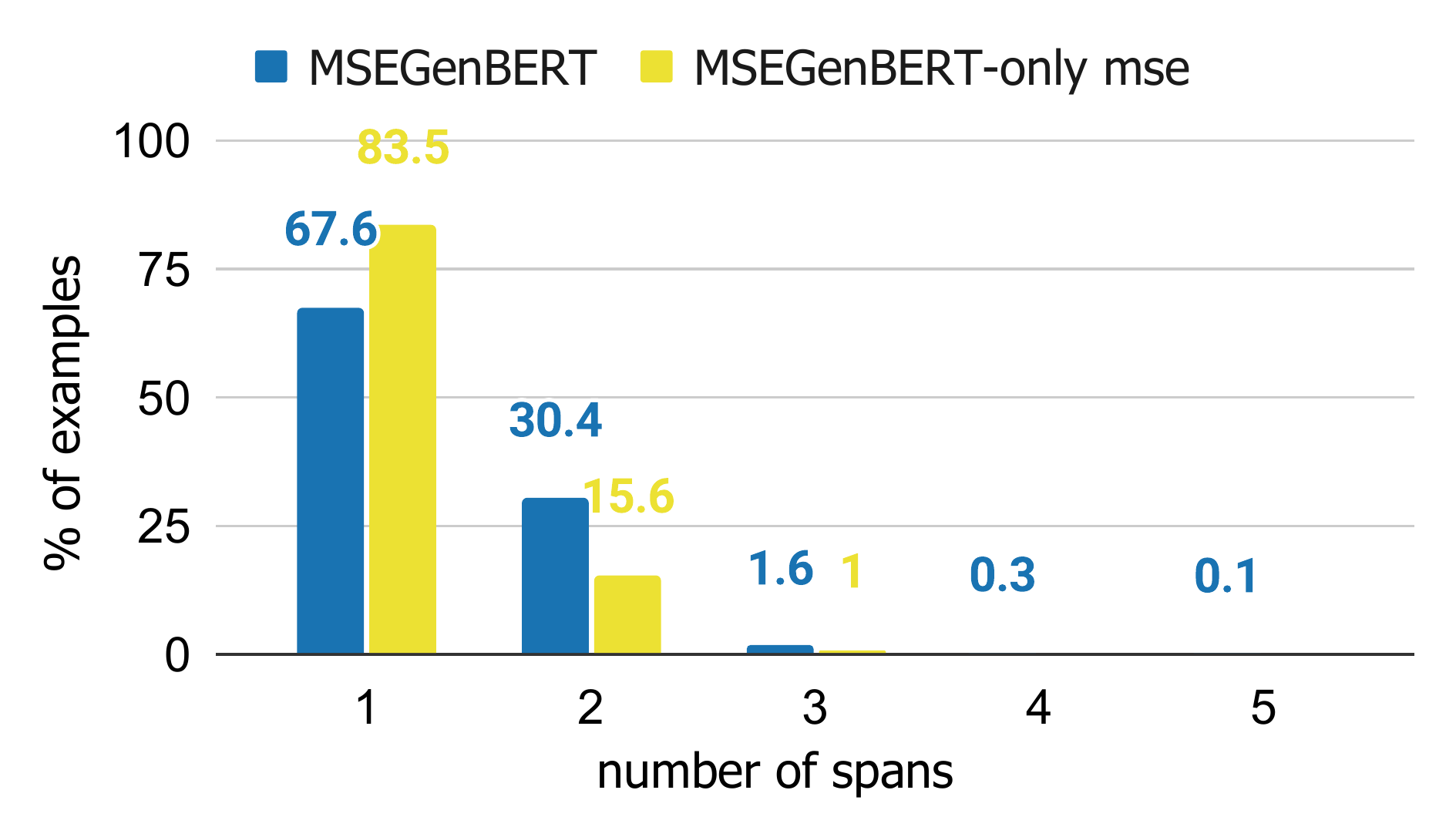}
    \caption{The portion of examples per number of extracted spans by \omse{}, for \msegenbert{} that was trained on with and without \ogen{}.}
    \label{figure:num_spans_mawps}
\end{figure}

To further test the emergent behaviour of \msegenbert{} (\S\ref{sec:numerical_arguments_extraction}), we compare the number of extracted spans on an out-of-distribution sample, by \msegenbert{} and \msegenbert{}$_{\text{only mse}}$ that was trained without the decoder head (\ogen{}).
Specifically, we run the models on \textsc{MAWPS} \cite{koncel2016mawps}, a collection of small-size math word problem datasets.
The results, shown in Figure~\ref{figure:num_spans_mawps}, demonstrate the generalized behaviour of \omse{}, which learns to extract multiple spans when trained jointly with the decoder.
 
\section{Example Emergent Query-based Summaries}
\label{sec:example_emergent_summaries}

Examples are provided in Tables~\ref{table:example_summary_1},~\ref{table:example_summary_2},~\ref{table:example_summary_3}, and~\ref{table:example_summary_4}.

\begin{table}[t]
    \centering
    \footnotesize
    \begin{tabular}{p{7.3cm}}
          \hline
          \emph{Question}: Who is Bruce Spizer an expert on, known as the most influential act of the rock era? (\nl{The Beatles}) \\
          \emph{Context}: \textbf{The Beatles were an English rock band formed in Liverpool in 1960.} \textbf{With members John Lennon, Paul McCartney, George Harrison and Ringo Starr, they became widely regarded as the foremost and most influential act of the rock era.} Rooted in skiffle, beat and 1950s rock and roll, the Beatles later experimented with several musical styles, ranging from pop ballads and Indian music to psychedelia and hard rock, often incorporating classical elements and unconventional recording techniques in innovative ways. In 1963 their enormous popularity first emerged as ``Beatlemania'', and as the group's music grew in sophistication in subsequent years, led by primary songwriters Lennon and McCartney, they came to be perceived as an embodiment of the ideals shared by the counterculture of the 1960s. \textbf{David ``Bruce'' Spizer (born July 2, 1955) is a tax attorney in New Orleans, Louisiana, who is also recognized as an expert on The Beatles.} He has published eight books, and is frequently quoted as an authority on the history of the band and its recordings. \\ \hline
    \end{tabular}
    \caption{Example (1) input from \hotpotqa{} and the predicted summary by \readersum{}. The summary is marked in bold.}
    \label{table:example_summary_1}
\end{table}

\begin{table}[t]
    \centering
    \footnotesize
    \begin{tabular}{p{7.3cm}}
          \hline
          \emph{Question}: Which Eminem album included vocals from a singer who had an album titled ``Unapologetic''? (\nl{The Marshall Mathers LP 2}) \\
          \emph{Context}: \textbf{``Numb'' is a song by Barbadian singer Rihanna from her seventh studio album ``Unapologetic'' (2012).} \textbf{It features guest vocals by American rapper Eminem, making it the pair's third collaboration since the two official versions of ``Love the Way You Lie''.} Following the album's release, ``Numb'' charted on multiple charts worldwide including in Canada, the United Kingdom and the United States. \textbf{``The Monster'' is a song by American rapper Eminem, featuring guest vocals from Barbadian singer Rihanna, taken from Eminem's album ``The Marshall Mathers LP 2'' (2013).} The song was written by Eminem, Jon Bellion, and Bebe Rexha, with production handled by Frequency. ``The Monster'' marks the fourth collaboration between Eminem and Rihanna, following ``Love the Way You Lie'', its sequel ``Love the Way You Lie (Part II)'' (2010), and ``Numb'' (2012). ``The Monster'' was released on October 29, 2013, as the fourth single from the album. The song's lyrics present Rihanna coming to grips with her inner demons, while Eminem ponders the negative effects of his fame. \\ \hline
    \end{tabular}
    \caption{Example (2) input from \hotpotqa{} and the predicted summary by \readersum{}. The summary is marked in bold.}
    \label{table:example_summary_2}
\end{table}

\begin{table}[t]
    \centering
    \footnotesize
    \begin{tabular}{p{7.3cm}}
          \hline
          \emph{Question}:Are both Dictyosperma, and Huernia described as a genus? (\nl{yes}) \\
          \emph{Context}:\textbf{The genus Huernia (family Apocynaceae, subfamily Asclepiadoideae) consists of stem succulents from Eastern and Southern Africa, first described as a genus in 1810}. The flowers are five-lobed, usually somewhat more funnel- or bell-shaped than in the closely related genus "Stapelia", and often striped vividly in contrasting colours or tones, some glossy, others matt and wrinkled depending on the species concerned. To pollinate, the flowers attract flies by emitting a scent similar to that of carrion. \textbf{The genus is considered close to the genera "Stapelia" and "Hoodia"}. The name is in honour of Justin Heurnius (1587–1652) a Dutch missionary who is reputed to have been the first collector of South African Cape plants. His name was actually mis-spelt by the collector.\textbf{Dictyosperma is a monotypic genus of flowering plant in the palm family found in the Mascarene Islands in the Indian Ocean (Mauritius, Reunion and Rodrigues)}. The sole species, Dictyosperma album, is widely cultivated in the tropics but has been farmed to near extinction in its native habitat. It is commonly called princess palm or hurricane palm, the latter owing to its ability to withstand strong winds by easily shedding leaves. It is closely related to, and resembles, palms in the "Archontophoenix" genus. The genus is named from two Greek words meaning "net" and "seed" and the epithet is Latin for "white", the common color of the crownshaft at the top of the trunk. \\ \hline
    \end{tabular}
    \caption{Example (3) input from \hotpotqa{} and the predicted summary by \readersum{}. The summary is marked in bold.}
    \label{table:example_summary_3}
\end{table}

\begin{table}[t]
    \centering
    \footnotesize
    \begin{tabular}{p{7.3cm}}
          \hline
          \emph{Question}: Which board game was published most recently, Pirate's Cove or Catan? (\nl{Pirate's Cove'}) \\
              \emph{Context}:\textbf{The Settlers of Catan, sometimes shortened to Catan or Settlers, is a multiplayer board game designed by Klaus Teuber and first published in 1995 in Germany by Franckh-Kosmos Verlag (Kosmos) as Die Siedler von Catan}. Players assume the roles of settlers, each attempting to build and develop holdings while trading and acquiring resources. Players are awarded points as their settlements grow; the first to reach a set number of points, typically 10, is the winner. \textbf{The game and its many expansions are also published by Mayfair Games, Filosofia, Capcom, 999 Games, $\mathbf{K\alpha\iota\sigma\sigma\alpha}$, and Devir.}\textbf{Pirate's Cove (in German, Piratenbucht) is a board game designed by Paul Randles and Daniel Stahl, originally published in Germany in 2002 by Amigo Spiele, illustrated by Markus Wagner and Swen Papenbrock}. In 2003, Days of Wonder republished the game with a new graphic design from Julien Delval and Cyrille Daujean. In the game, players play pirate ship captains seeking treasure from islands and bragging rights from defeating other pirates in naval combat. \\ \hline
    \end{tabular}
    \caption{Example (4) input from \hotpotqa{} and the predicted summary by \readersum{}. The summary is marked in bold.}
    \label{table:example_summary_4}
\end{table}

\end{document}